\documentclass{article}
\usepackage{hyperref}
\hypersetup{
    colorlinks=true,
    linkcolor=red,
    filecolor=green,      
    urlcolor=blue,
    citecolor=green,
}
\usepackage{spconf,amsmath,graphicx}
\usepackage{cite}
\usepackage{graphicx}
\usepackage{textcomp}
\usepackage{xcolor}
\usepackage{booktabs}  
\usepackage{color}     
\usepackage{colortbl}  
\usepackage{hyperref}
\usepackage{subcaption}
\usepackage{graphicx}
\usepackage{algorithm}
\usepackage{algpseudocode}
\usepackage{bbding} 
\usepackage{multirow} 
\newcommand{\smallerfonttable}{\fontsize{6pt}{7pt}\selectfont}

\title{LRDif: Diffusion Models for Under-Display Camera Emotion Recognition}
\name{Zhifeng Wang, Kaihao Zhang, Ramesh Sankaranarayana}
\address{School of Computing, Australian National University, Canberra, Australia, 2601}
\begin{document}
%
\maketitle
\begin{abstract}
This study introduces LRDif, a novel diffusion-based framework designed specifically for facial expression recognition (FER) within the context of under-display cameras (UDC). To address the inherent challenges posed by UDC's image degradation, such as reduced sharpness and increased noise, LRDif employs a two-stage training strategy  that integrates a condensed  preliminary extraction network (FPEN) and an agile transformer network (UDCformer) to effectively identify emotion labels from UDC images. By harnessing the robust distribution mapping capabilities of Diffusion Models (DMs) and the spatial dependency modeling strength of transformers, LRDif effectively overcomes the obstacles of noise and distortion inherent in UDC environments. Comprehensive experiments on standard FER datasets including RAF-DB, KDEF, and FERPlus, LRDif demonstrate state-of-the-art performance, underscoring its potential in advancing FER applications. This work not only addresses a significant gap in the literature by tackling the UDC challenge in FER but also sets a new benchmark for future research in the field.
\end{abstract}
\section{Introduction}
\begin{figure}[t]
\centering
\includegraphics[width=0.95\linewidth]{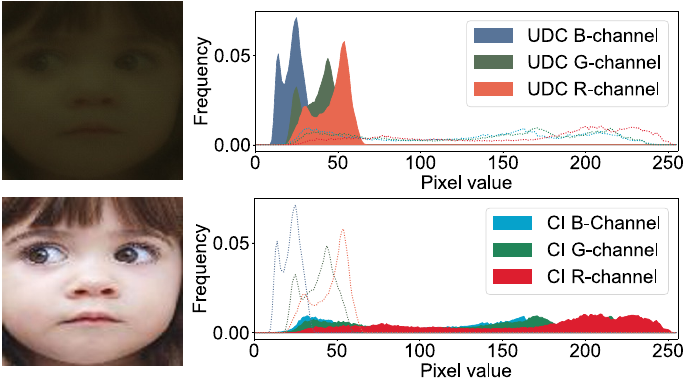}
\caption{Top: shows an image taken with an under-display camera (UDC), which looks less clear compared to a regular camera. Bottom: shows an image taken with a traditional external camera, which is much clearer. }
\label{udc_clear_images_histogram}
\end{figure}

Facial expression recognition (FER) has undergone remarkable advancements in recent years. However, implementing emotion recognition under UDC environment poses unique challenges. The fundamental challenge lies in the image quality and clarity. UDC images often suffer from reduced sharpness, increased noise, and color fidelity issues compared to those captured with traditional external cameras in Fig .\ref{udc_clear_images_histogram}. These quality constraints stem from the fact that the camera lens is positioned beneath the screen, which can obstruct light in unpredictable ways. For emotion recognition algorithms, which rely heavily on the nuances of facial expressions, this can lead to decreased accuracy.

Moreover, UDC images may exhibit unique artifacts and lighting inconsistencies, further complicating the task. Machine learning models used for emotion recognition need to be adapted for UDC images, ensuring they can effectively recognise emotional states despite the additional noise and distortions. Previous studies on FER \cite{2021arm,2021dmue,2021manet,zhang2017facial,niu2022four,wang2023htnet,zhang2015facial} have not sufficiently focused on the impacts of additional noise and distortions brought by UDC images. However, tackling this issue is essential for improving the real-world application of full-screen devices.

Currently, several approaches address the noise learning problem in the emotion recognition field. RUL \cite{2021rul} addresses uncertainties due to ambiguous expressions and inconsistent labels by weighting facial features based on their relative difficulty, improving performance in noisy data environments. 
EAC \cite{2022eac} addresses noisy labels by utilizing flipped semantic consistency and selective erasure of input images, which prevents the model from overly focusing on specific features associated with noisy labels. SCN \cite{2020scn} mitigates uncertainties in large-scale datasets by employing a self-attention mechanism to weight training samples and a relabeling mechanism to adjust the labels of low-ranked samples, reducing overfitting to uncertain facial images. However, these methods encounter limitations when applied to UDC images. Specifically, RUL \cite{2021rul} and EAC \cite{2022eac} rely on the assumption of small losses, which can lead to confusion between difficult and noisy samples since both tend to exhibit high loss values during training. Therefore, learning features from noisy labels and images remains a challenging task.

\begin{figure*}[t]
\centering
\includegraphics[width=0.74\linewidth]{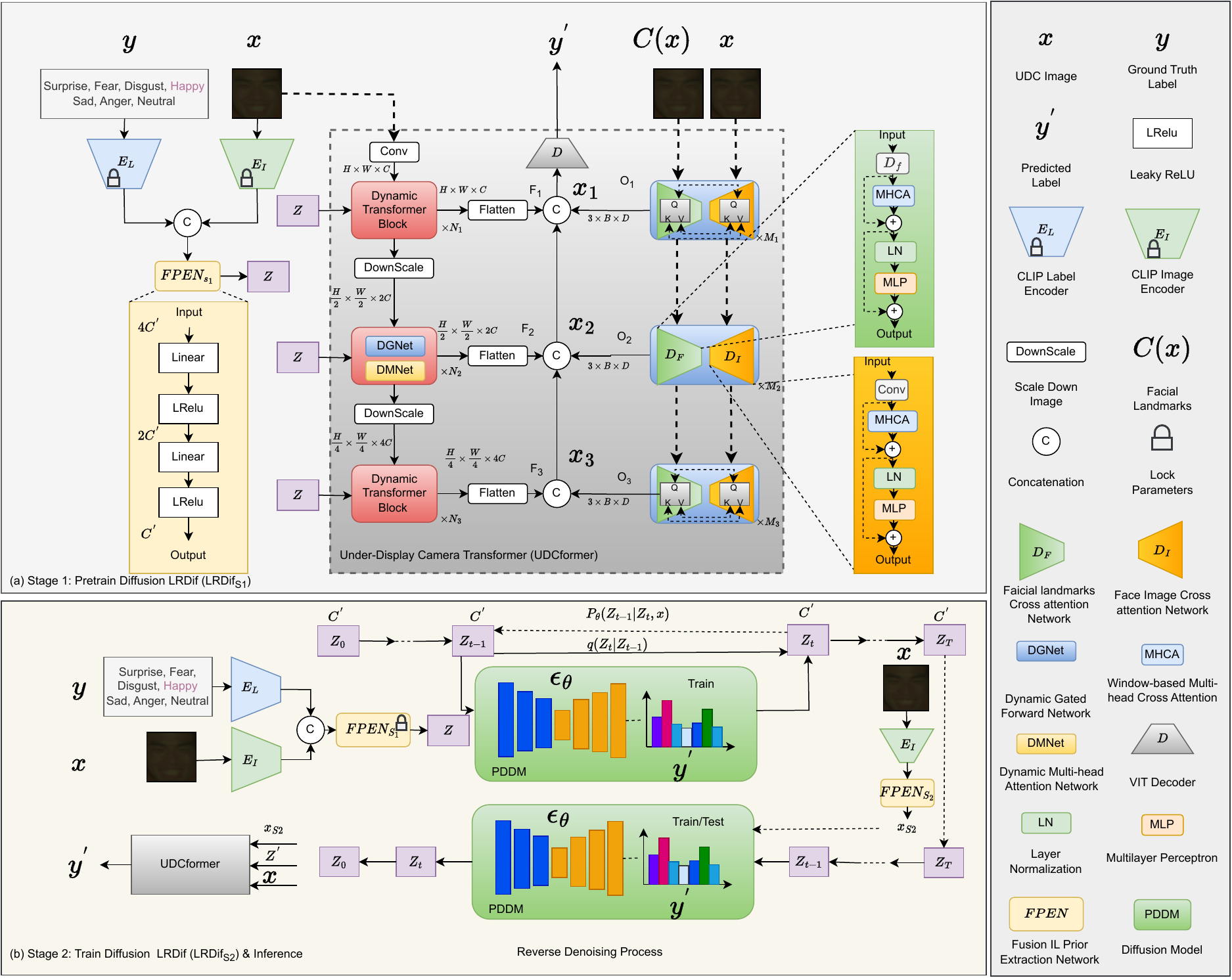}
\caption{The proposed LRDif framework, comprising UDCformer, FPEN, and a denoising network. LRDif has two distinct training stages:(a) In stage 1, $FPEN_{S_1}$ processes the  UDC image and ground-truth emotion label to generate an EPR Z, which guides the UDCformer on label restoration. (b) In stage 2, the robust data prediction capabilities of the DM are employed to assess the EPR produced by the pre-trained $FPEN{S_1}$. It's important to note that during the inference phase, the model relies solely on the DM's reverse process, omitting the need for actual labels.}
\label{LRDif_architecture}
\end{figure*}

To address this issue, this paper adopts a fresh perspective on noisy UDC images. Rather than the traditional approach of identifying noisy samples based on their loss values, this paper introduces an innovative view that focuses on learning from noisy labels through feature extraction. Our goal is to create a diffusion-based FER system that utilises the distribution mapping capabilities of diffusion models (DMs) to effectively restore noise labels and their corresponding images. For this purpose, we introduce LRDif. Considering the transformer's proficiency in capturing long-distance pixel dependencies, we choose transformer blocks as the foundational components of our LRDif architecture. Dynamic transformer blocks are layered in a U-Net style to construct the Under-Display Camera Transformer (UDCformer), aimed at multi-level feature extraction. The UDCformer comprises two parallel networks: the DTnetwork, which extracts latent features from label and UDC images at multiple levels, and the DILnetwork, which learns the similarities between facial landmarks and UDC images. LRDif follows a two-stage training process:  (1) Initially, as shown in Fig. \ref{LRDif_architecture} (a), we construct a condensed  preliminary extraction network (FPEN) that extracts an Emotion Prior Representation (EPR) from latent label and UDC images. This EPR is then utilized to guide the UDCformer in restoring labels. (2) Subsequently, shown in Fig. \ref{LRDif_architecture} (b), the diffusion model (DM) is trained to directly infer the precise EPR from UDC images. Owing to the lightweight nature of EPR $Z$, the DM can achieve highly accurate EPR predictions, leading to significant improvements in test accuracy after a few iterations. 

The main contributions of this work are summarized as follows: 1) We introduce a novel approach to address the challenges posed by under-display cameras (UDC) in facial expression recognition, offering a diffusion-based solution aimed at mitigating the effects of extra noise and image distortions. 2) LRDif focuses on the powerful mapping capabilities of DMs to deduce a concise emotion prior representation (EPR), enhancing both the accuracy and consistency of FER predictions. This method distinguishes itself from previous approaches by not relying on the knowledge of the dataset's uncertainty distribution. 3) Extensive experiments demonstrate that LRDif achieves state-of-the-art performance in emotion recognition tasks on three synthesized UDC-FER datasets and several standard FER datasets.

\section{Related Work}

\textbf{Facial Expression Recognition}. A typical FER system comprises three main phases: face detection, feature extraction, and expression classification. In the face detection phase, tools such as MTCNN \cite{mtcnn} and Dlib \cite{dlip} are used to identify faces, which may then be aligned for further processing. During the feature extraction phase, various techniques \cite{2023posterv2} are employed to capture the distinct facial features associated with different expressions. Wang \textit{et al.} \cite{wang2021light} utilize a spatial attention mechanism to concentrate on emotion-relevant areas of the image, thus enhancing accuracy in real-world conditions with a lightweight network that can be integrated into standard convolutional neural networks. MRAN \cite{chen2023multi} improves performance in uncontrolled conditions by applying spatial attention to both global and local facial features and using transformers to understand the relationships within and between these features.

\textbf{Diffusion Models}. Diffusion Models are applied in various fields such as image editing \cite{yang2023paint}, image restoration \cite{wang2023lldiffusion,chen2023towards}, and high-resolution image synthesis \cite{zhang2022styleswin}. Rombach \textit{et al.} \cite{rombach2022high} enhance the efficiency and visual fidelity of diffusion models by training them in the latent space of pre-trained autoencoders. They integrate cross-attention layers for versatile conditioning inputs, enabling high-resolution image synthesis with reduced computational demands while preserving detail. DiffusionDet \cite{chen2023diffusiondet} introduces an innovative object detection framework that conceptualizes the detection process as a denoising diffusion from random to precise object boxes. It learns to reverse the diffusion process from the ground truth during training and iteratively refines randomly generated boxes during inference.

\section{Methods}
\subsection{Pretrained DTnetwork}
 There are two basic networks in the stage 1: the condensed  preliminary extraction network (FPEN) and the agile transformer network (DTnetwork). Illustrated in the yellow box of Fig. \ref{LRDif_architecture}, FPEN includes several linear layers to extract the EPR.  After that, the DTnetwork utilizes this EPR to recover labels. The DTnetwork's architecture, illustrated in the red box of Fig. \ref{LRDif_architecture}, comprises the Dynamic Multi-Head Transposed Attention (DMNet) and the Gated Feed-Forward Network (DGNet). During the pretraining phase, illustrated in Fig. \ref{LRDif_architecture} (a), both FPEN$_{S1}$ and the DTnetwork are trained together. We employ CLIP \cite{radford2021clip} text and image encoders to obtain latent features from labels and UDC images, which are then fed into FPEN$_{S1}$. The output of FPEN$_{S1}$ is the EPR $Z \in R^C$. This process is depicted in Equation (\ref{Z_FPEN_S1}):
\begin{equation}
   Z = FPEN_{S1}(Conc(E_L(y),E_I(x))).
   \label{Z_FPEN_S1}
\end{equation}

Then, $Z$ is input into the DGNet and DMNet within the DTnetwork, serving as learnable parameters to aid in label recovery, as described in (Eq. (\ref{F_w1_ln})).
\begin{equation}
F^{'} = W_{1}^{l}Z\circ LN(F) +  W_{2}^{l}Z,
\label{F_w1_ln}
\end{equation}
where,  LN refers to  layer normalization, $W$ is the weight of fully connected layer, $\circ$ stands for element-wise multiplication. 

Next, in the DMNet, we extract information from the entire image. We convert the processed features, $F^{'}$, into three new vectors called query $Q$, key $K$ and value $V$ by using convolution layer. Then, we reshape $Q$ as $R^{H^{"}W^{"}\times C^{"}}$, $K$ as $R^{C^{"}\times H^{"}W^{"}}$, and $V$ as $R^{H^{"}W^{"}\times C^{"}}$ into new formats suitable for further calculations. We multiply $Q$ and $K$, which helps our model understand which parts of the image to focus on, creating attention map $A \in R^{C^{"}\times C^{"}}$. This whole step in DMNet can be summarized as follows (Eq. (\ref{f_wc_v_softmax})):
\begin{equation}
F^{"} = W_{c}V\times softmax(K\times Q/\alpha) + F,
\label{f_wc_v_softmax}
\end{equation}
where $\alpha$ is an adjustable parameter during training. Next,  the DGNet learns local and adjacent features through aggregation. We use a very small Conv ($1\times 1$) to merge details from different layers and a slightly larger  Conv ($3\times 3$) that looks at each layer separately to gather details from nearby pixels. In addition, we use a special gate to ensure the capture of the most useful information. This entire step in DGNet can be summarised as (Eq. (\ref{f_gelu_w1_w1c})):
\begin{equation}
F^{"} = GELU(W^{1}_{d}W^{1}_{c}F^{'})\circ W^{2}_{d}W^{2}_{c}F^{'} + F.
\label{f_gelu_w1_w1c}
\end{equation}

\subsection{Dynamic Image and Landmarks Network (DILnetwork)}
In the DILnetwork, we employ a window-based cross-attention mechanism to extract features from facial landmarks and UDC images. For UDC image features, represented as $X_{udc} \in R^{N\times D}$, we initially segment them into several distinct, non-overlapping segments $x_{udc}\in R^{M\times D}$. As for the facial landmarks features, denoted by $X_{flm} \in R^{C\times H\times W}$, we scale them down to match the size of these segments, resulting in $x_{flm} \in R^{c\times h\times w}$, where the dimension $c$ is equal to $D$ and the product of $h$ and $w$ equals $M$. This allows us to perform cross-attention between the facial landmarks and UDC image features using $N$ attention heads, as detailed in (Eq. (\ref{Q_O_O_FLM})).
\begin{align}
Q   &=x_{flm}w_{Q}, K = x_{udc}w_{K}, V = x_{udc}w_{V},\\
O_{i}&= Softmax(\frac{Q_{i}K_{i}^T}{\sqrt{d}} +b)V_{i}, i=1,...,N,\\
O &= [O_{1},O_{2},...,O_{N}]W_{O},
\label{Q_O_O_FLM}
\end{align}
where $w_{Q}$ $w_{K}$, $w_{V}$, $w_{O}$ are weight matrix of features and $b$ is the related position bias.

This cross-attention is applied to each window within the UDC image, which refers as window-based multi-head cross-attention (MHCA). The LRDif's transformer encoder is described by the following equations (Eq. (\ref{x-udc-mhca})):
\begin{align}
X^{'}_{udc} &= MHCA(X_{udc}) +X_{udc},\\
X^{"}_{udc} &= MLP(LN(X^{'}_{udc})) + X^{'}_{udc},
\label{x-udc-mhca}
\end{align}

We need to merge the output features $F$ from DTnetwork with the output features $O$ from DILnetwork to get the fused multi-scale features $x_1$, $x_2$ and $x_3$, where  $x_1 = Concat(F_1,O_1)$, $x_2 =Concat(F_2,O_2)$ and $x_3 =Concat(F_3,O_3)$. Then, these fused features $X$ will input into vanilla transformer blocks for  further processing.
\begin{align}
X &=[x_1,x_2,x_3],\\
X^{'} &=MSA(X) +X,\\
y^{'} &=MLP(LN(X^{'})) +X^{'},
\end{align}
where $MSA$ is multi-head self-attention blocks. $LN$ is layer normalization function.The training loss is defined as follows (Eq. (\ref{l_ce_loss})):
\begin{equation}
\mathcal{L}_{ce} = - \sum_{i=1}^{N} \sum_{c=1}^{M} y_{ic} \log(p_{ic}).
\label{l_ce_loss}
\end{equation}
We use cross-entropy loss to train our model, with N representing the total number of samples and M the total number of classes. In this context, $y_{ic}$ denotes the presence or absence of class c as the correct label for sample i (1 if correct, 0 if not), while $p_{ic}$ represents the model's predicted probability that sample i falls into class c.

\subsection{Diffusion Model for Label Restoration}
 In stage 2 (Fig. \ref{LRDif_architecture} (b)), we utilise the powerful capabilities of DM to estimate the Emotion Prior Representation. Initially, we utilize the pre-trained FPEN$_{S1}$ to obtain the EPR $Z \in R^{C}$. Subsequently, we apply the diffusion process to $Z$ to generate a sample $Z_T \in R^{C}$, as detailed in Equation (\ref{sample_z}):
\begin{equation}
q(Z_T | Z) = \mathcal{N}(Z_T; \sqrt{\bar{\alpha}_T} Z, (1 - \bar{\alpha}_T)I).
\label{sample_z}
\end{equation}
In this context, $T$ denotes the total number of diffusion steps, with $\alpha_{t} = 1-\beta_{t}$ and $\bar{\alpha}_T$ being the cumulative product of $\alpha_i$ from 0 to $T$. The term $\beta_{t}$ refers to a predefined hyper-parameter, and $\mathcal{N}(.)$ represents the standard Gaussian distribution.

In the reverse process of DM, we initially utilize the  CLIP image encoder $E_I$ to encode the input UDC image $x$ (Eq. (\ref{x_s2_fpen})). Subsequently, the encoded features are input into FPEN$_{S2}$ to abtain a conditional vector $x_{S2} \in R^{C}$ from the UDC images. 
\begin{equation}
x_{S2} = \text{FPEN}_{S2}(E_{I}(x)),
\label{x_s2_fpen}
\end{equation}
where $\text{FPEN}_{S2}$ shares similar network structures as $\text{FPEN}_{S1}$. Only the first layer's dimension is different.
The denoising network, denoted by $\epsilon_{\theta}$, estimates the noise at each time step $t$. It takes as input the current noisy data $Z_{t}^{'}$, the time step $t$, and a conditional vector $x_{S2} $, derived from the UDC image through the second-stage condensed preliminary extraction network FPEN$_{S2}$. The estimated noise, given by $\epsilon_{\theta}(\text{Concat}(Z_{t}^{'},t,x_{S2}))$, is then used in the following equation to calculate the denoised data $Z_{t-1}^{'}$  for the next iteration (Eq. (\ref{reverse_noise_process})) :
\begin{equation}
Z_{t-1}^{'}  = \frac{1}{\sqrt{\alpha_t}}(Z_{t}^{'} - \epsilon_{\theta}(\text{Concat}(Z_{t}^{'},t,x_{S2}))\frac{1-\alpha_{t}}{\sqrt{1-\alpha_{t}}}).
\label{reverse_noise_process}
\end{equation}
After a series of $T$ iterations, the final estimated Emotion Prior Representation (EPR), denoted by $Z_0'$, is obtained. The condensed  preliminary extraction network for stage two (FPEN$_{S2}$), the denoising network, and the Under-Display Camera Transformer (UDCformer) are jointly trained using the total loss function $\mathcal{L}_{total}$ (Eq. (\ref{l_total_loss})).
\begin{equation}
\mathcal{L}_{kl} = \sum_{i=1}^{C} Z_{\text{norm}}(i) \log(\frac{Z_{\text{norm}}(i)}{\bar{Z}_{\text{norm}}(i)}),
\label{l_kl_loss}
\end{equation}
\begin{equation}
\mathcal{L}_{total} = \mathcal{L}_{ce} + \mathcal{L}_{kl} ,
\label{l_total_loss}
\end{equation}
where $Z_{\text{norm}}(i)$ and $\bar{Z}_{\text{norm}}(i)$ are EPRs extracted by LRDif$_{S1}$ and LRDif$_{S2}$, respectively, both of which are normalized using the softmax operation. $\mathcal{L}_{kl}$ is a variant of the Kullback-Leibler divergences, calculated across C dimensions. We combine the Kullback-Leibler divergence loss $\mathcal{L}_{kl}$ (Eq. \ref{l_kl_loss}) and the Cross-Entropy loss $\mathcal{L}_{ce}$ (Eq. \ref{l_ce_loss}) to compute the total loss $\mathcal{L}_{total}$ (Eq. \ref{l_total_loss}).  Given that the Emotion Prior Representation (EPR) includes both the under-display camera (UDC) image features and the associated emotion label encoded via CLIP, the LRDif stage two (LRDif$_{S2}$) is able to achieve reliable estimations within a limited number of iterations. During the inference phase, LRDif does not use the ground truth labels in the reverse diffusion process.
\begin{table*}[t]
\centering
\smallerfonttable 
\renewcommand{\arraystretch}{0.7} 
\caption{Comparison results with SOTA FER algorithms on RAF-DB, FERPlus and KDEF.}
\label{three-datasets-sota-comparison}
\begin{tabular}{lc|lc|lcc}
\toprule 
RAF-DB &   & FERPlus &  & KDEF &  \\
\midrule 
Methods & Acc. (\%) & Methods & Acc. (\%)  & Methods & Acc. (\%) \\
\midrule 
ARM\cite{2021arm}     & 90.42  &DACL\cite{2021dacl}  & 83.52 & DACL\cite{2021dacl} &88.61  \\
POSTER++\cite{2023posterv2} & \textcolor{blue}{92.21} & POSTER++\cite{2023posterv2} & \textcolor{blue}{86.46} & POSTER++\cite{2023posterv2} & \textcolor{blue}{94.44} \\
RUL\cite{2021rul}     & 88.98  & RUL\cite{2021rul}  & 85.00  & RUL\cite{2021rul} & 87.83\\
DAN\cite{2023dan}     & 89.70  & DAN\cite{2023dan}  & 85.48       & DAN\cite{2023dan} &  88.77\\
SCN\cite{2020scn}     & 87.03  & SCN\cite{2020scn}  & 83.11  & SCN\cite{2020scn} & 89.55 \\
EAC\cite{2022eac}     & 90.35  & EAC\cite{2022eac}  & 86.18  & EAC\cite{2022eac} & 72.32\\
MANet\cite{2021manet} & 88.42  & MANet\cite{2021manet}&85.49 &MANet\cite{2021manet} & 91.75 \\
\midrule 
$Ours$     & \textcolor{red}{92.24} &$Ours$   &\textcolor{red}{87.13}  &$Ours$   & \textcolor{red}{95.75}\\
\bottomrule 
\end{tabular}
\end{table*}
\begin{table}[t]
\centering
\smallerfonttable 
\setlength{\tabcolsep}{1pt} 
\caption{Comparison of accuracy (\%) with state-of-the-art results on the UDC-RAF-DB Dataset.}
\label{udc-raf-db-sota-comparison}
\begin{tabular}{lcccccccc}
\toprule 
\parbox{0pt}{\rule{0pt}{5ex}} & ARM\cite{2021arm} & POSTER++\cite{2023posterv2} & RUL\cite{2021rul} & DAN\cite{2023dan} & SCN\cite{2020scn} &EAC\cite{2022eac} & MANet\cite{2021manet} &$Ours$ \\
\midrule 
 Acc.(\%)  & 86.44 & \textcolor{blue}{86.76} & 85.59 & 86.47 & 85.89 & 86.51&85.62 & \textcolor{red}{88.55}\\
\bottomrule 
\end{tabular}
\end{table}
\begin{figure}[tp]
\centering
\includegraphics[width=0.9\linewidth]{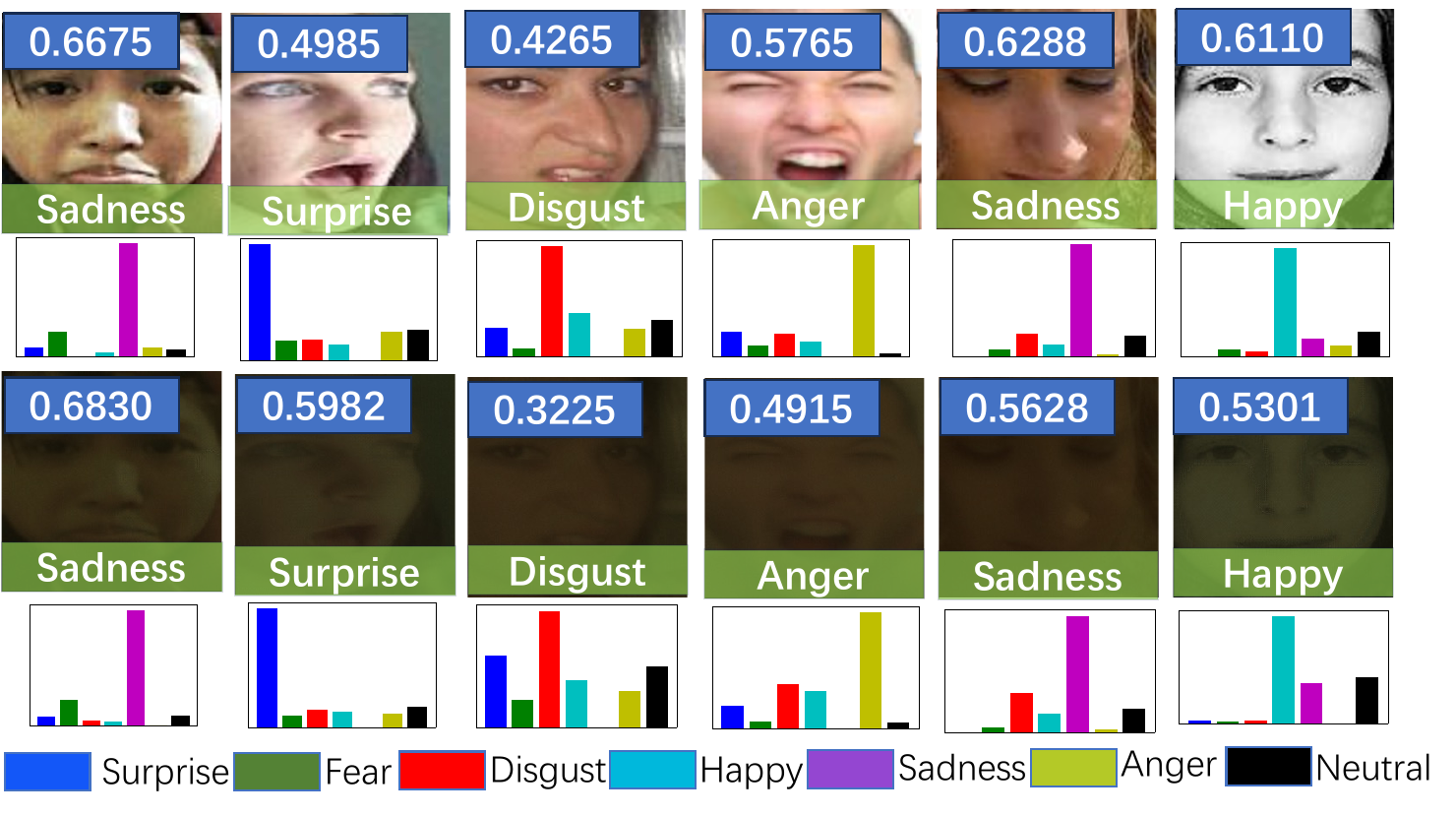}
\caption{Samples from RAF-DB and UDC-RAF-DB datasets}
\label{probability-distribution-ours}
\end{figure}
\begin{table}[t]
\centering
\smallerfonttable 
\setlength{\tabcolsep}{1pt} 
\renewcommand{\arraystretch}{1} 
\caption{Comparison of accuracy (\%) with state-of-the-art results on the UDC-FERPlus Dataset.}
\label{udc-ferplus-sota-comparison}
\begin{tabular}{lcccccccc}
\toprule 
\parbox{0pt}{\rule{0pt}{5ex}} & DACL\cite{2021dacl}  & POSTER++\cite{2023posterv2} & RUL\cite{2021rul} & DAN\cite{2023dan} & SCN\cite{2020scn} &EAC\cite{2022eac} & MANet\cite{2021manet} &$Ours$ \\
\midrule 
 Acc.(\%)  & 78.11 & \textcolor{blue}{83.78} & 82.15 & 83.25  & 77.38 & 82.72&83.19 & \textcolor{red}{84.89}\\
 
\bottomrule 
\end{tabular}
\end{table}
\begin{table}[ht]
\centering
\smallerfonttable 
\setlength{\tabcolsep}{1pt} 
\renewcommand{\arraystretch}{1} 
\caption{Comparison of accuracy (\%) with state-of-the-art results on the UDC-KDEF Dataset.}
\label{udc-kdef-sota-comparison}
\begin{tabular}{lcccccccc}
\toprule 
\parbox{0pt}{\rule{0pt}{5ex}} & DACL\cite{2021dacl}  & POSTER++\cite{2023posterv2} & RUL\cite{2021rul} & DAN\cite{2023dan} & SCN\cite{2020scn} &EAC\cite{2022eac} & MANet\cite{2021manet} &$Ours$ \\
\midrule 
 Acc.(\%)  & 84.44 &  \textcolor{blue}{91.92} &  82.69  &  85.71  & 78.69 & 54.31 &88.48 & \textcolor{red}{94.07}\\
\bottomrule 
\end{tabular}
\end{table}
\section{Experiments}
\subsection{Datasets}
\textbf{UDC-RAF-DB} dataset comprises a training set with 12,271 images and a testing set encompassing 3,068 images,providing a solid foundation for developing and evaluating FER algorithms in UDC environments. Similarly, the RAF-DB dataset \cite{rafdb} encompasses seven emotional categories and shares the same training and testing configurations as the UDC-RAF-DB dataset. Both datasets feature a nearly identical distribution of expressions.

\textbf{UDC-FERPlus} dataset expands the UDC domain, offering a comprehensive set of 28,709 UDC images for training and 7,178 for testing. FERPlus \cite{ferplus} builds upon the FER2013 dataset, augmented with a new set of labels provided by ten annotators. This enhanced dataset comprises 28,709 images for training and 7,178 for testing.

\textbf{UDC-KDEF} dataset includes a total of 4,900 UDC images, offering varied perspectives with images captured from five different angles. The training set comprises 3,920 UDC images, while the testing set includes 980 images. KDEF \cite{kdef} represents a comprehensive dataset encompassing a collection of 4,900 images with same settings. 

\subsection{Implementation Details}
 We use the SOTA MPGNet \cite{zhou2022modular} to synthesize three benchmark UDC-FER datasets. The image degradation process contains brightness attenuation, blurring, and noise corruption. The experiments were conducted using PyTorch, and the models were trained on a GTX-3090 GPU. We utilized the Adam optimizer for training over 100 epochs. The training protocol involved a batch size of 64, a learning rate of $3.5 \times 10^{-4}$, and a weight decay of $1 \times 10^{-4}$.
\subsection{Comparison with SOTA FER Methods}
\textbf{Comparison with Typical FER-model.} Table \ref{three-datasets-sota-comparison} presents a comprehensive comparison of accuracy  between the proposed method and current state-of-the-art (SOTA) FER algorithms \cite{2020scn,2021manet,2021arm,2022eac,2021dacl,2021rul,2023dan,2023posterv2}  across three benchmark datasets: RAF-DB, FERPlus, and KDEF. For the RAF-DB dataset, the proposed method (`Ours') achieves an accuracy of 92.24\%, surpassing several established algorithms such as ARM\cite{2021arm}, RUL\cite{2021rul}, DAN\cite{2023dan}, SCN\cite{2020scn}, EAC\cite{2022eac}, and MANet\cite{2021manet}, and is competitive with POSTER++\cite{2023posterv2}, which scores slightly lower at 92.21\%. In the FERPlus dataset, the proposed method also demonstrates superior performance with an accuracy of 87.13\%, outperforming other notable methods like DACL\cite{2021dacl}, RUL\cite{2021rul}, DAN\cite{2023dan}, SCN\cite{2020scn}, and MANet\cite{2021manet}. Remarkably, in the KDEF dataset comparison, the proposed method achieves the highest accuracy of 95.75\%, indicating a significant advancement over other methodologies, including the second-highest performing POSTER++\cite{2023posterv2} at 94.44\% and the lower-scoring MANet\cite{2021manet} at 91.75\%. Overall, this table underscores the robustness and effectiveness of the proposed method in facial expression recognition tasks, as evidenced by its leading accuracy on diverse datasets.
\begin{figure}[t] 
\centering
\begin{subfigure}{0.45\linewidth}
\includegraphics[width=\linewidth,height = 1.8cm]{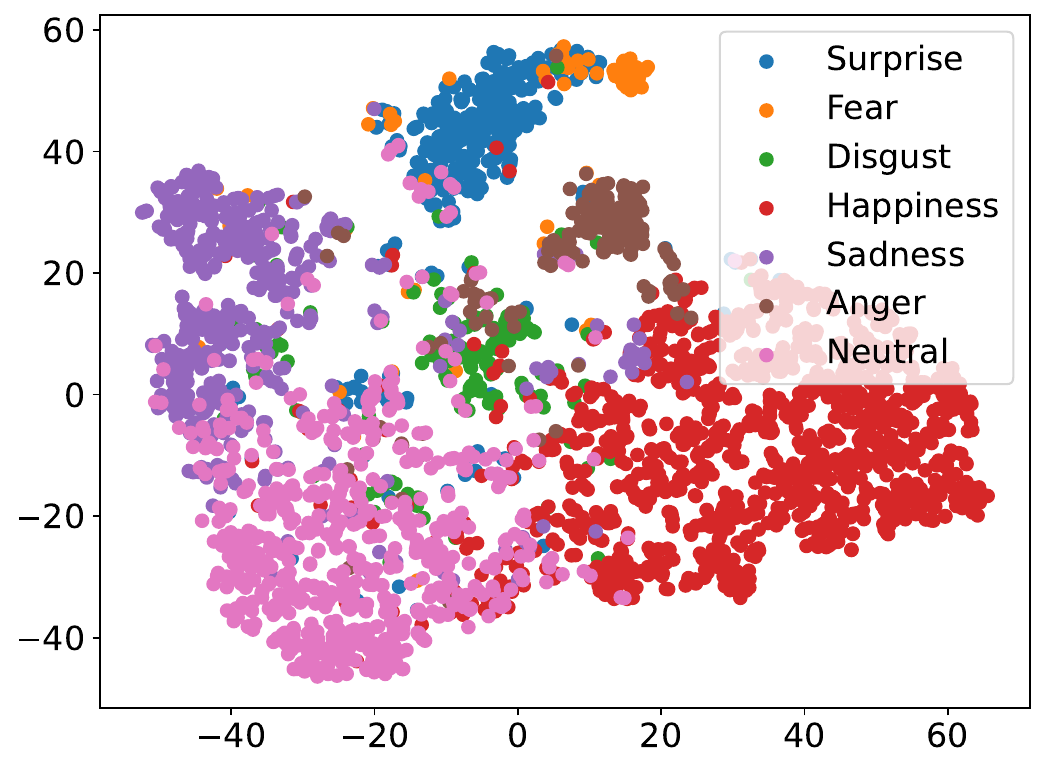}
\caption{SCN on clear images }
\end{subfigure}
\begin{subfigure}{0.45\linewidth}
\includegraphics[width=\linewidth,height = 1.8cm]{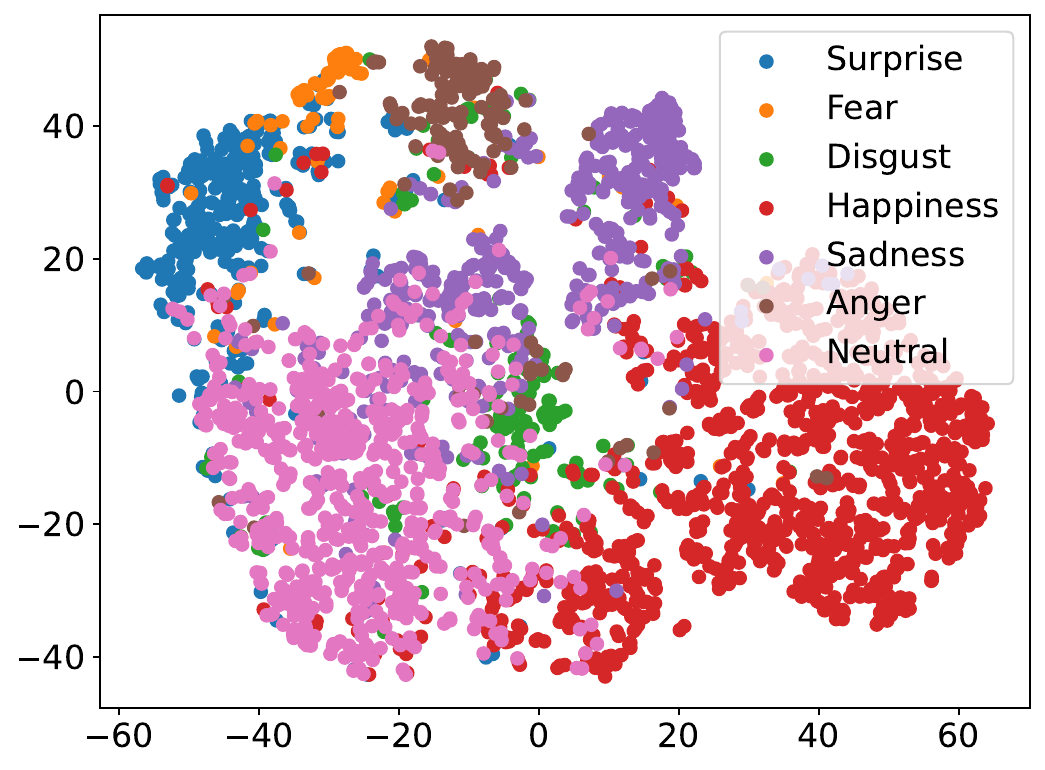}
\caption{SCN on UDC images  }
\end{subfigure}
\begin{subfigure}{0.45\linewidth}
\includegraphics[width=\linewidth,height = 1.8cm]{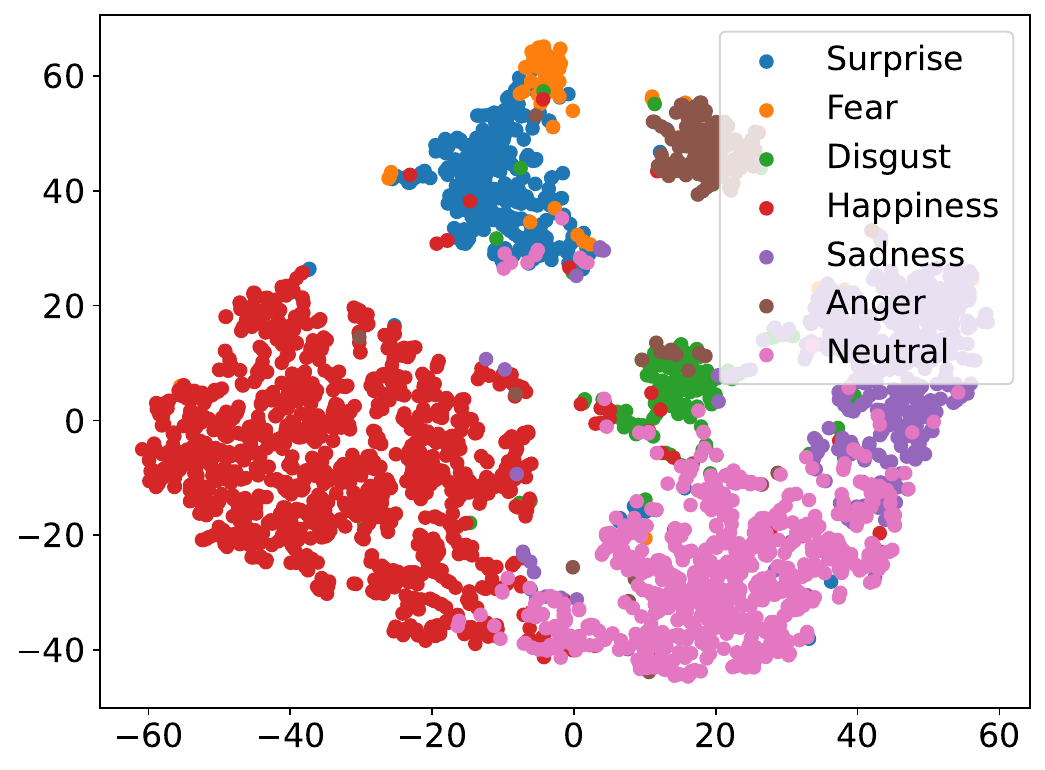}
\caption{LRDif on clear images }
\end{subfigure}
\begin{subfigure}{0.45\linewidth}
\includegraphics[width=\linewidth,height = 1.8cm]{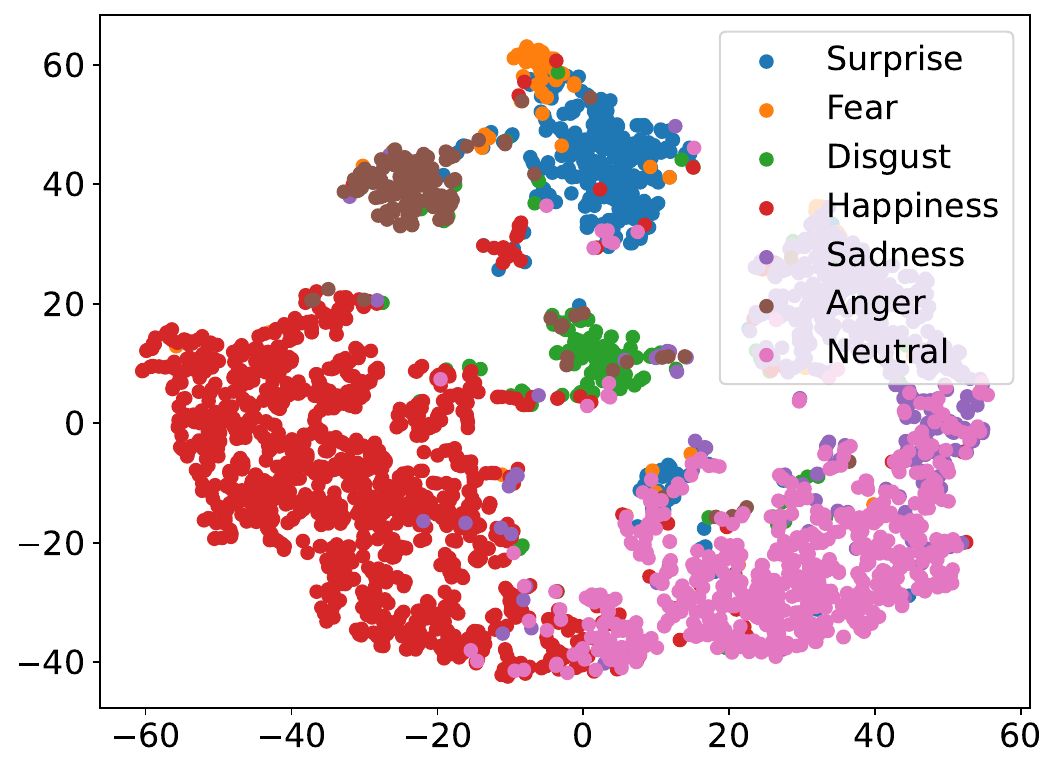}
\caption{LRDif on UDC images  }
\end{subfigure}
\caption{The learned feature distribution by SCN and LRDif training on the RAF-DB datasets.}
\label{tsne_scn_ours_comparision}
\end{figure}

\textbf{Comparison with the UDC FER-model.} We conducted a performance evaluation of our proposed model for UDC systems, with some samples illustrated in Fig. \ref{probability-distribution-ours}. Table \ref{udc-raf-db-sota-comparison}, Table \ref{udc-ferplus-sota-comparison} and  Table \ref{udc-kdef-sota-comparison} present a comparative analysis of accuracy among various state-of-the-art FER models on the UDC-RAF-DB, UDC-FERPlus, and UDC-KDEF datasets, respectively.  Models such as ARM\cite{2021arm}, RUL\cite{2021rul}, DAN\cite{2023dan}, SCN\cite{2020scn}, EAC\cite{2022eac}, and MANet\cite{2021manet} primarily employ the ResNet-18 architecture. An exception is POSTER++\cite{2023posterv2}, which utilizes the Vision Transformer architecture, and ARM\cite{2021arm}, which uses a modified ResNet-18 architecture termed ResNet18-ARM. Our proposed model deviates from the conventional ResNet-18 framework by integrating a `Diffusion' backbone. This novel approach achieves an accuracy of 88.55\%, 84.89\% and 94.07\% on three UDC datasets, indicating a significant improvement in performance compared to other methods. The results highlight the effectiveness of the diffusion-based model in the realm of UDC-based FER systems.
\begin{table}[tpb]
\centering
\smallerfonttable 
\setlength{\tabcolsep}{0.8pt} 
\renewcommand{\arraystretch}{0.4} 
\caption{Accuracy (\%) on the UDC-RAF-DB Dataset.}
\label{ablation_study_components}
\renewcommand{\arraystretch}{1} 
\begin{tabular}{l|cccccccc}  
\toprule[1.2pt] 
\multirow{2}{*}{\textbf{Method}} & \multicolumn{5}{c}{\textbf{Core components in LRDif}} &\multirow{2}{*}{\textbf{Acc.(\%)}} \\ \cmidrule(l){2-6} 
&  \textbf{Ground Truth} & \textbf{Diffusion Model} &\textbf{$\mathcal{L}_{ce}$} &\textbf{$\mathcal{L}_{total}$} & \textbf{Insert Noise} & \\
\midrule 
LRDif$_{S1}$  &  \CheckmarkBold & \XSolidBrush & \CheckmarkBold& \XSolidBrush & \XSolidBrush & 100\\
\midrule 
LRDif$_{S2}$-V1  &  \XSolidBrush &  \XSolidBrush  & \CheckmarkBold& \XSolidBrush& \XSolidBrush & 86.08\\
LRDif$_{S2}$-V2  &\XSolidBrush & \CheckmarkBold  & \CheckmarkBold & \XSolidBrush& \XSolidBrush & 88.98\\
LRDif$_{S2}$-V3 (Ours) &\XSolidBrush& \CheckmarkBold  &  \XSolidBrush &  \CheckmarkBold & \CheckmarkBold & 88.55 \\
LRDif$_{S2}$-V4  & \XSolidBrush & \CheckmarkBold  & \XSolidBrush & \CheckmarkBold & \XSolidBrush & 88.78\\
\bottomrule[1.2pt] 
\end{tabular}
\end{table}

\textbf{Feature Visualization.} The t-SNE technique was employed to visualize the feature distribution patterns learned by different models. Unlike in Fig. \ref{tsne_scn_ours_comparision} (a) and (b), where the SCN model struggles to distinguish between emotion categories, particularly in UDC images, our LRDif model demonstrates effective recognition of expressions within mixed and noisy images. This suggests that LRDif successfully identifies the most distinctive features essential for differentiating various emotional expressions.
\subsection{Ablation Study}
In this section, we evaluate the impact of key components of LRDif such as the Diffusion Model (DM), the loss functions, and the incorporation of noise during training, as presented in Table \ref{ablation_study_components}. (1) The comparison between LRDif$_{S2}$-V3 and LRDif$_{S2}$-V1 highlights the DM's robust capability for precise EPR prediction. (2) For LRDif$_{S2}$-V4, removing noise in the DM process is shown to  enhance the precision of EPR estimations. (3) Furthermore, we evaluate the effectiveness of loss functions. We find that employing LRDif$_{S2}$-V2 with $\mathcal{L}_{ce}$ (as per Eq. (\ref{l_ce_loss})) and LRDif$_{S2}$-V4 with $\mathcal{L}_{total}$ (as per Eq. (\ref{l_total_loss})), indicates that $\mathcal{L}_{ce}$ contributes to enhanced accuracy.
\begin{figure}[t] 
\centering
\begin{subfigure}{0.31\linewidth}
\caption{T = 1 }
\includegraphics[width=\linewidth]{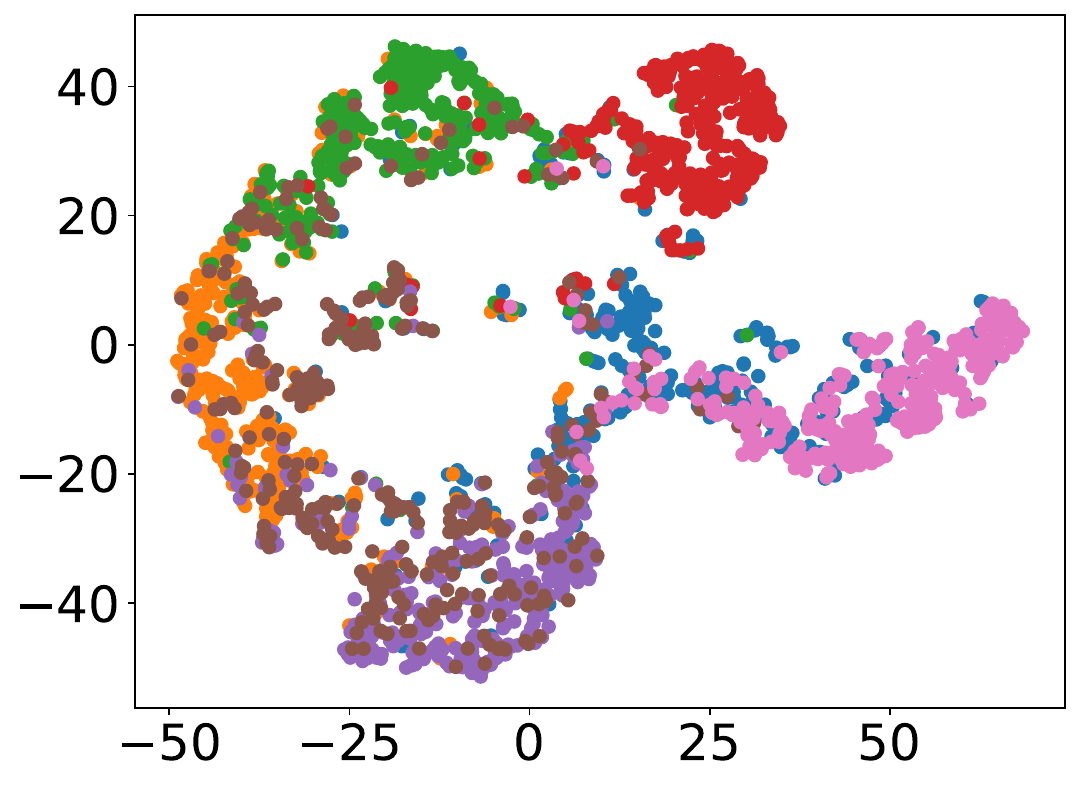}
\end{subfigure}
\begin{subfigure}{0.31\linewidth}
\caption{T = 2 }
\includegraphics[width=\linewidth]{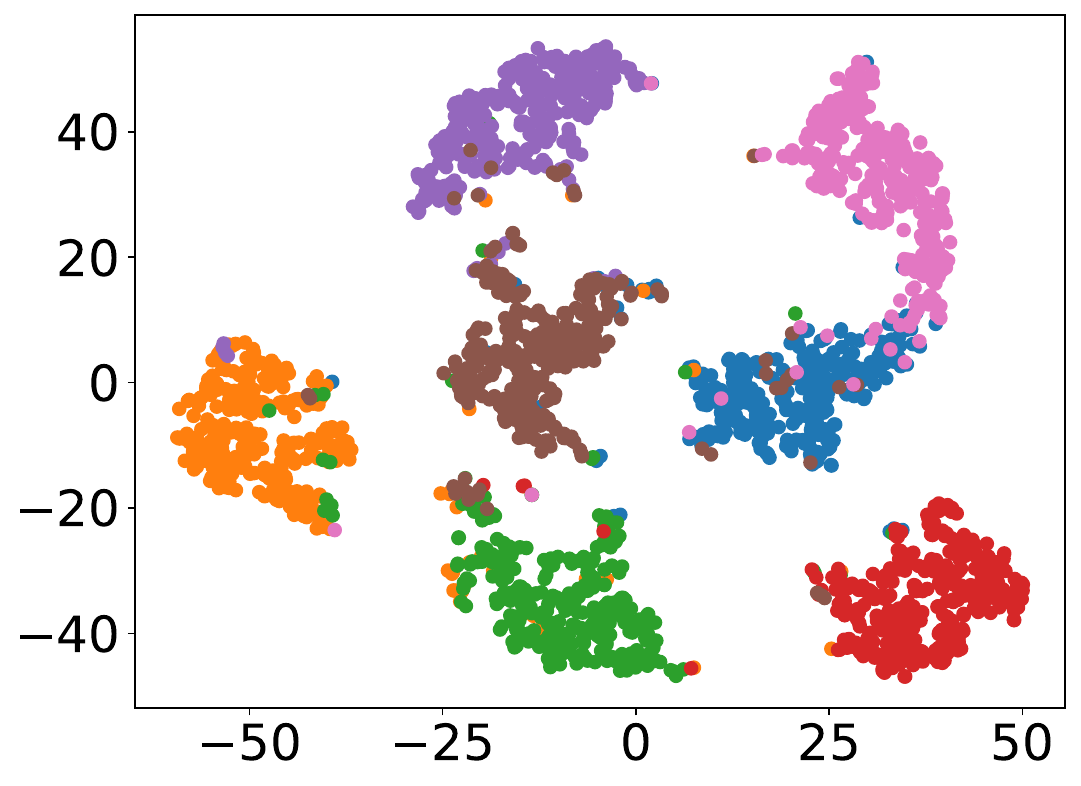}
\end{subfigure}
\begin{subfigure}{0.31\linewidth}
\caption{T = 4}
\includegraphics[width=\linewidth]{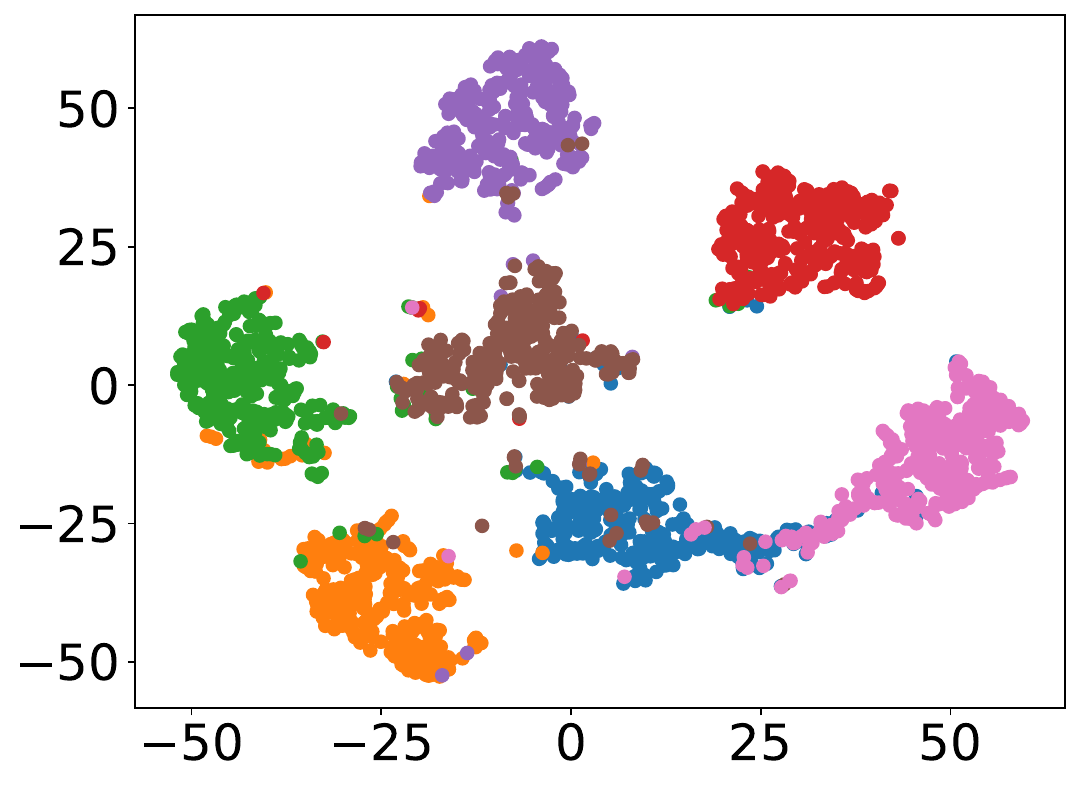}
\end{subfigure}
\begin{subfigure}{0.31\linewidth}
\caption{T= 8}
\includegraphics[width=\linewidth]{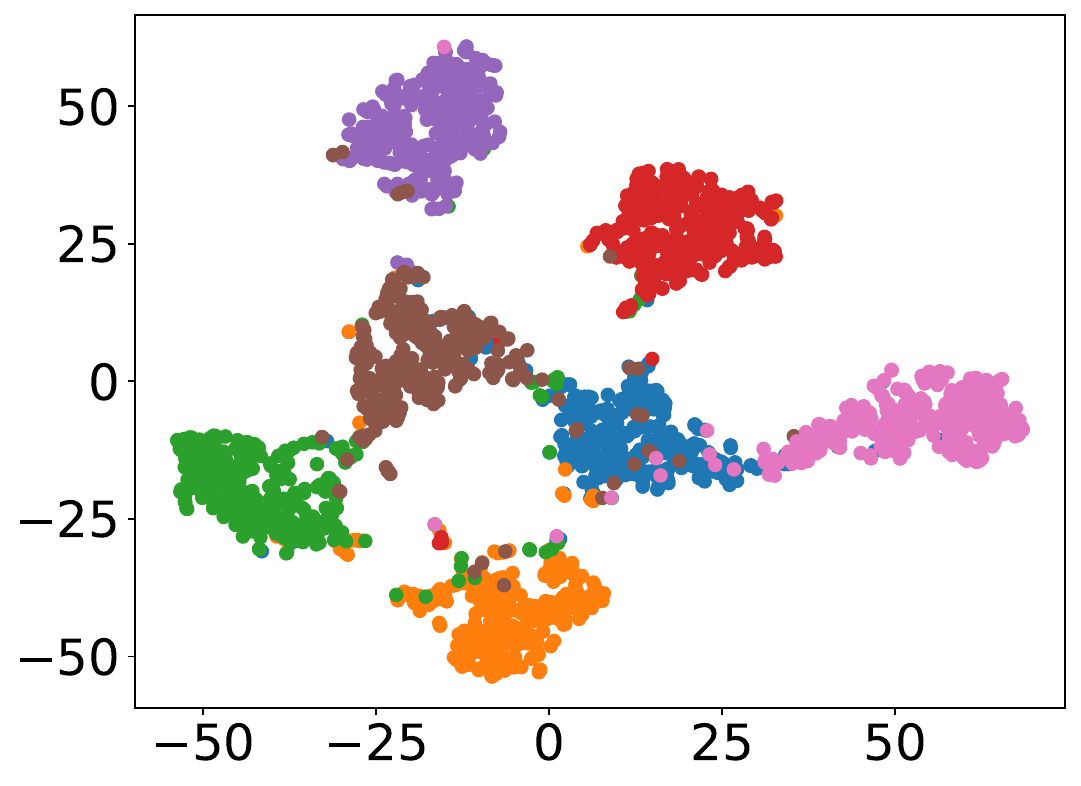}
\end{subfigure}
\begin{subfigure}{0.31\linewidth}
\caption{T= 16}
\includegraphics[width=\linewidth]{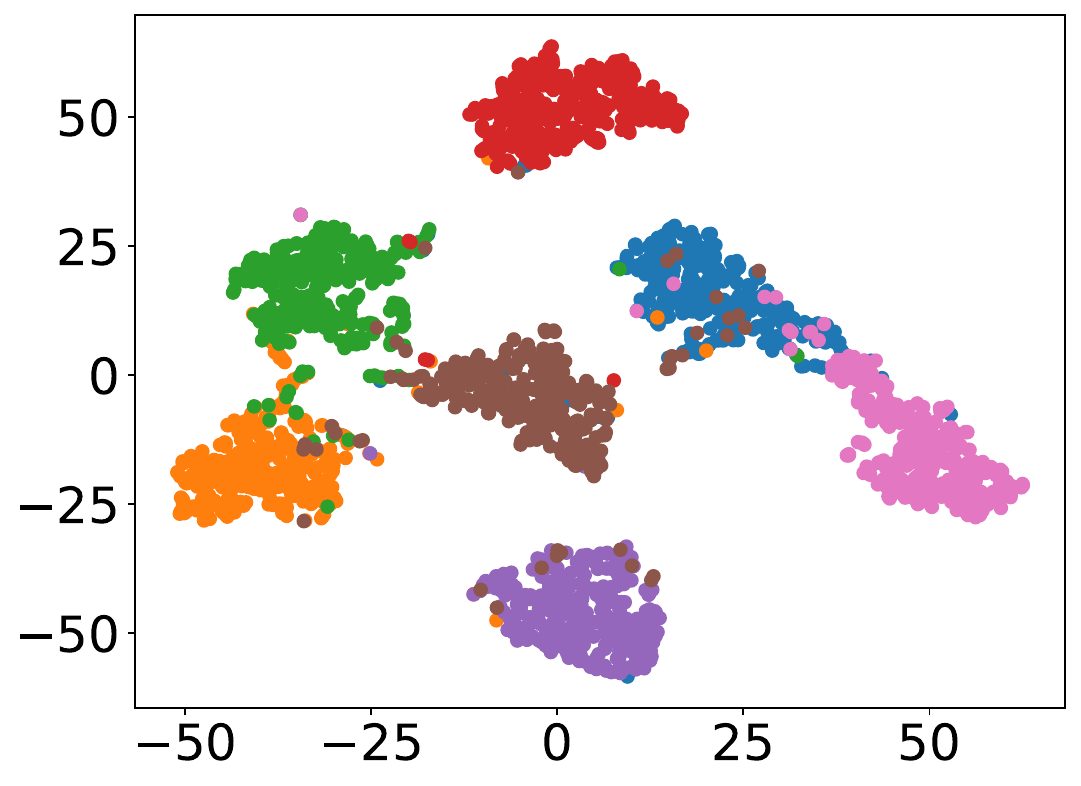}
\end{subfigure}
\begin{subfigure}{0.31\linewidth}
\caption{T= 32}
\includegraphics[width=\linewidth]{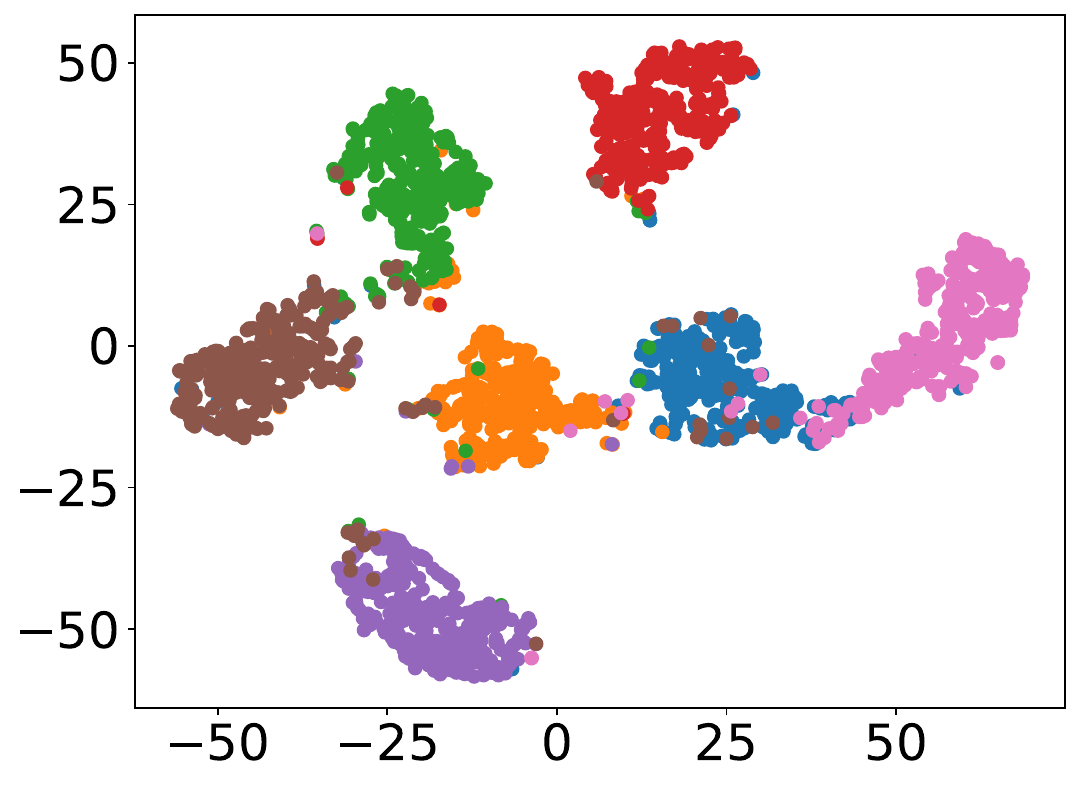}
\end{subfigure}
\caption{t-SNE feature visualization on for DM trained on UDC-KDEF datasets.}
\label{t-sne-feature-DDPM}
\end{figure}

\begin{figure}[t]
\centering
\includegraphics[width=0.9\linewidth]{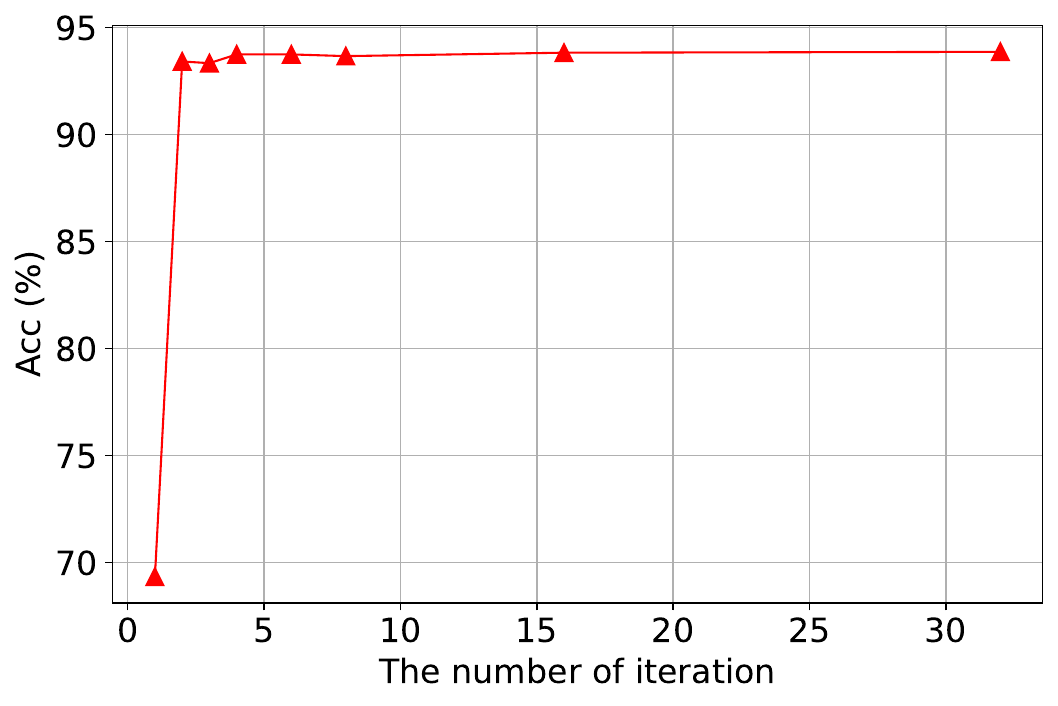}
\caption{Investigation of number of iterations in DM.}
\label{number_of_iterations}
\end{figure}
\textbf{Impact of iteration numbers.} 
This section explores the effect of the number of iterations within the Diffusion Model (DM) on the performance of LRDif$_{S2}$. We tested various iteration counts in LRDif$_{S2}$, adjusting the $\beta_t$ parameter (with $\alpha_t = 1 - \beta_t$, as defined in Eq. \ref{sample_z}) to ensure that the variable $Z$ converges to Gaussian noise, $Z_T \sim \mathcal{N}(0,1)$. As illustrated in Figs. \ref{number_of_iterations} and \ref{t-sne-feature-DDPM}, LRDif$_{S2}$ shows a significant performance improvement when the iteration count reaches 4. Beyond this point, further increases in the number of iterations do not markedly affect performance, indicating an optimal threshold has been reached. Notably, LRDif${S2}$ achieves faster convergence compared to traditional DM approaches, which often require more than 50 iterations. This increased efficiency is attributed to the application of the DM to the EPR, a concise, one-dimensional vector.
\section{Conclusion}


This paper introduces LRDif, a novel diffusion-based framework for facial expression recognition in UDC environments. LRDif overcomes UDC image degradation through a two-stage training strategy, integrating a preliminary extraction network (FPEN) and a transformer network (UDCformer). These modules enable effective recovery of emotion labels from degraded UDC images. 
%
%
Experimental results indicate that the proposed DRDif model demonstrates superior performance, setting a new benchmark in state-of-the-art results across three UDC facial expression datasets.

\bibliographystyle{IEEEbib}
\bibliography{refs}

\end{document}